\documentclass{article}
\usepackage{spconf,amsmath,graphicx,tabularx,epstopdf,url}

\def\w{\mathbf{w}}
\def\x{\mathbf{x}}
\def\z{\mathbf{z}}

\def\bmu{\boldsymbol{\mu}}
\def\R{\mathcal{R}}
\def\I{\mathbf{I}}
\def\W{\mathbf{W}}
\def\X{\mathbf{X}}
\def\Z{\mathbf{Z}}
\def\Sbar{\bar{\mathbf{S}}}
\def\Xbar{\bar{\mathbf{X}}}
\def\SBu{\mathbf{S}_{B_U}}
\def\SWu{\mathbf{S}_{W_U}}
\def\SBp{\mathbf{S}_{B_P}}

\DeclareMathOperator{\tr}{tr}

\newcolumntype{Y}{>{\centering\arraybackslash}X}

\title{Ratio Utility and Cost Analysis for Privacy Preserving Subspace Projection}
\name{Mert Al, Shibiao Wan, Sun-Yuan Kung\sthanks{This material is based upon work
supported in part by the Brandeis Program of the Defense
Advanced Research Project Agency (DARPA) and Space and
Naval Warfare System Center Pacific (SSC Pacific) under
Contract No. 66001-15-C-4068.}}
\address{Princeton University\\
Department of Electrical Engineering\\
Princeton, NJ, 08544, USA}
\begin{document}
\ninept
\maketitle
\begin{abstract}
With a rapidly increasing number of devices connected to the internet, big data has been applied to various domains of human life. Nevertheless, it has also opened new venues for breaching users' privacy. Hence it is highly required to develop techniques that enable data owners to privatize their data while keeping it useful for intended applications. Existing methods, however, do not offer enough flexibility for controlling the utility-privacy trade-off and may incur unfavorable results when privacy requirements are high. To tackle these drawbacks, we propose a compressive-privacy based method, namely RUCA (Ratio Utility and Cost Analysis), which can not only maximize performance for a privacy-insensitive classification task but also minimize the ability of any classifier to infer private information from the data. Experimental results on Census and Human Activity Recognition data sets demonstrate that RUCA significantly outperforms existing privacy preserving data projection techniques for a wide range of privacy pricings.

\end{abstract}
\begin{keywords}
Compressive privacy, Subspace methods, Projection matrix, Principal/Discriminant component analysis
\end{keywords}

\parskip 0pt

\section{Introduction}
\label{sec:intro}
With our daily activities moving online, vast amounts of personal information are being collected, stored and shared across the internet, often without the data owner's knowledge. Even when the data owners trust data keepers such as Internet Service Providers and Statistics Bureaus to keep their personal information private, the data are often needed to be analyzed and released for Statistics, Commercial and Research purposes. This raises obvious concerns about the privacy of data contributors, as not only are the data vulnerable to inadvertent leakages, but also to malicious inference by other parties. Thus privacy-protection methods should be employed that allow data collectors and owners to control the types of information that can be inferred from their data.

Consider a scenario where mobile users upload their sensor readings to the cloud, which in turn trains a classifier that allows smartphones to identify their users from sensor readings in the background as in \cite{Shi}. This approach takes advantage of the vast storage and computation resources of the cloud. However, without proper processing the same data can be used to infer sensitive information about users, such as location, context and activities performed \cite{Christin}. This is especially alarming given the fact that private information about users may not only be inferred by the cloud but possibly by other users as well through classifiers, which may include training samples in them \cite{Lin}.

A number of approaches based on data projection and/or noise addition have been proposed to preserve the statistics of the data for machine learning applications, while making privacy-sensitive information unavailable. Additive noise based randomization was proposed in \cite{RAgrawal}, but was shown to be susceptible to reconstruction attacks using spectral properties of random noise and data \cite{HKargupta}. Liu et al. \cite{KunLiu} proposed projection of the data to a lower dimensional space via a Random Projection Matrix. Later, a more suitable system was proposed in \cite{BinLiu} for collaborative-learning, where the cloud trains a classifier with data from multiple users. Each user randomly generates a hidden Projection Matrix and adds variable levels of noise to projected samples before sending them to the cloud.

In \cite{SYKungDCA,SYKungDUCA}, Kung presented a supervised version of Principle Component Analysis (PCA) called Discriminant Component Analysis (DCA) in order to project the data into a lower dimensional space that maximizes the discriminant power as in Fisher Discriminant Analysis \cite{SYKungKernelBook}. The recent work of Diamantaras and Kung \cite{Kostas} inspired by this approach introduced another criterion called Multiclass Discriminant Ratio (MDR), and projects the data based on a pair of desirable and undesirable classification tasks.  
 
Dimension reduction through data projection removes both application-relevant and privacy-sensitive information from the data. DCA and MDR attempt to remove as little application-relevant information as possible by optimizing the projection subspace for the intended classification task. Yet they do not offer any flexibility for finding a favorable trade-off between utility and privacy. 

To address these problems, we propose a methodology called RUCA (Ratio Utility and Cost Analysis), which forms a bridge between DCA (utility driven projection) and MDR (privacy emphasized projection) and allows data owners to select a compromise between them. RUCA can be considered as a generalization of DCA and MDR, and it can also be extended to multiple privacy-sensitive classifications. Experimental results on Census and Human Activity Recognition data sets show that our methodology can provide better classification accuracies for the desired task while outperforming state-of-the-art privacy preserving data projection methods in terms of accuracies obtained from privacy-sensitive classifications.

Our methodology for privacy preservation is described in Section \ref{sec:method}, and it is formulated as the problem of maximizing separability of projected data for a desired classification task, while minimizing separability for undesirable classifications. We then present Generalized Eigenvalue Decomposition as a method for finding the optimal Projection Matrix that achieves this task. Our methods are tested on real data with possible utility and privacy classifications in Section \ref{sec:eval} and are compared with other projection based privacy protection methods. Finally, we conclude in Section \ref{sec:conc}.

\section{Methodology}
\label{sec:method}

\subsection{Problem Statement}
For simplicity, we shall assume that there is a single privacy-sensitive classification on the data, though it is straightforward to generalize this to the case where there are multiple privacy-sensitive classifications. We assume that the data of our concern is fully represented by a set of $N$ $M$-dimensional vectors $\{\x_1,\x_2,\cdots,\x_N\}$. For the desired classification, which we name utility classification, we have a set of labels $y_i$ associated with the vectors $\x_i$. For an undesirable classification, which we name privacy classification, we have a set of labels $s_i$ associated with the vectors $\x_i$. There are two or more classes for each classification task, i.e. $y_i \in 1,\cdots,L$, $s_i \in 1,\cdots,P$, where $L$ and $P$ are the numbers of utility and privacy classes, respectively.

Let $\W$ be an $M \times K$ projection matrix where $K<M$ and $\z_i=\W^T\x_i$ denote the projection of a vector $\x_i$ to a $K$-dimensional subspace. Let $\X$ denote the $M \times N$ matrix whose columns correspond to the data entries $\x_i$ and $\Z$ denote the $K \times N$ matrix whose columns correspond to the projected entries $\z_i$. Given $\X$, our problem is to find a matrix $\W$ such that given the projected data matrix $\Z=\W^T\X$: 
\begin{enumerate}
 \item A classifier can achieve similar performance on the task of finding the labels $\{y_1,y_2,\cdots,y_N\}$, compared to the case where the full data matrix $\X$ is given.
 \item Conversely, any classifier achieves poor performance, ideally as poor as random guessing, on the task of finding the set of labels $\{s_1,s_2,\cdots,s_N\}$.
\end{enumerate}

\subsection{Projection Method}

\begin{figure*}[t]
\begin{minipage}[b]{0.33\linewidth}
  \centering
  \centerline{\includegraphics[width=6cm]{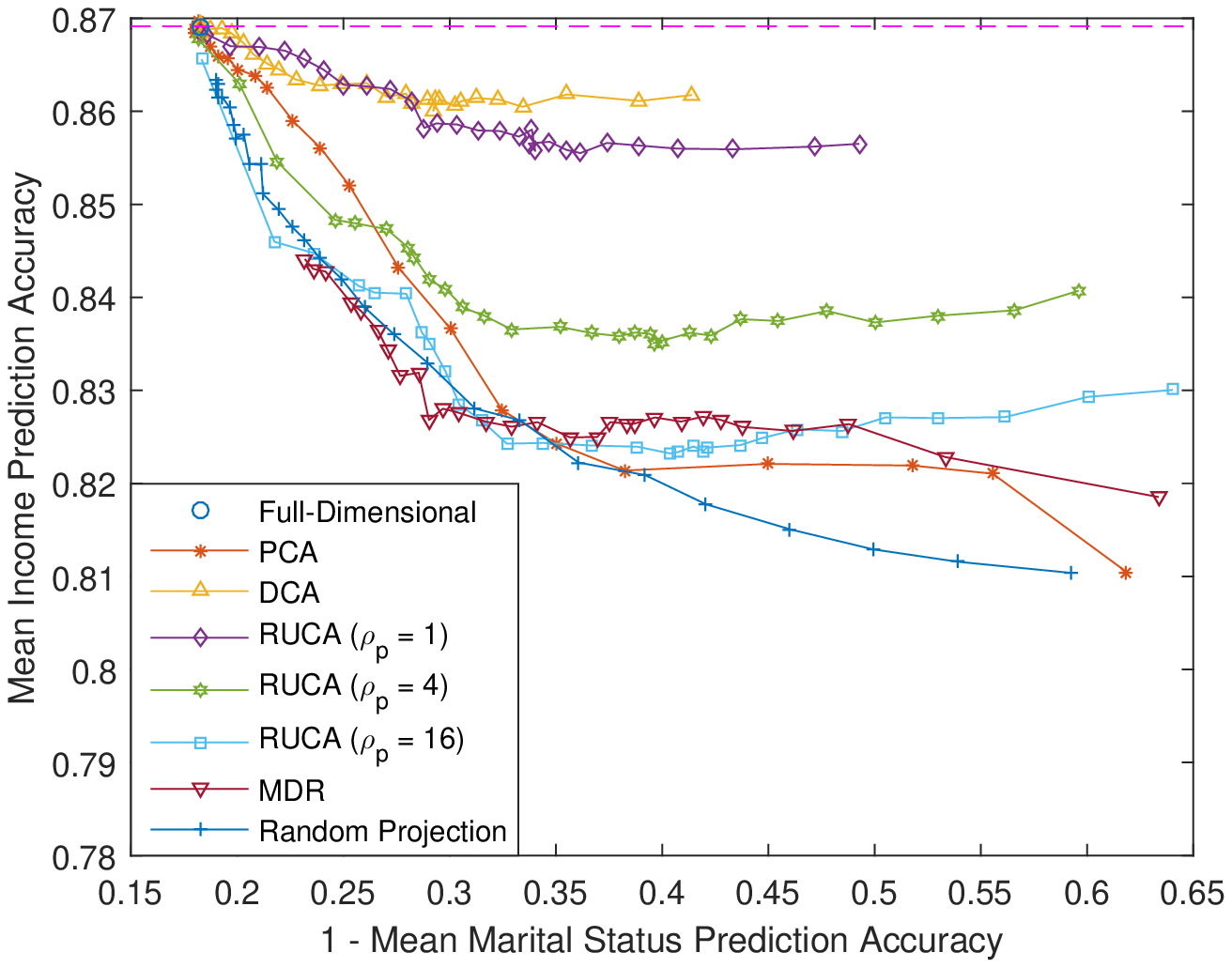}}
  \centerline{(a) Census}\medskip
\end{minipage}
\hfill
\begin{minipage}[b]{0.33\linewidth}
  \centering
  \centerline{\includegraphics[width=6cm]{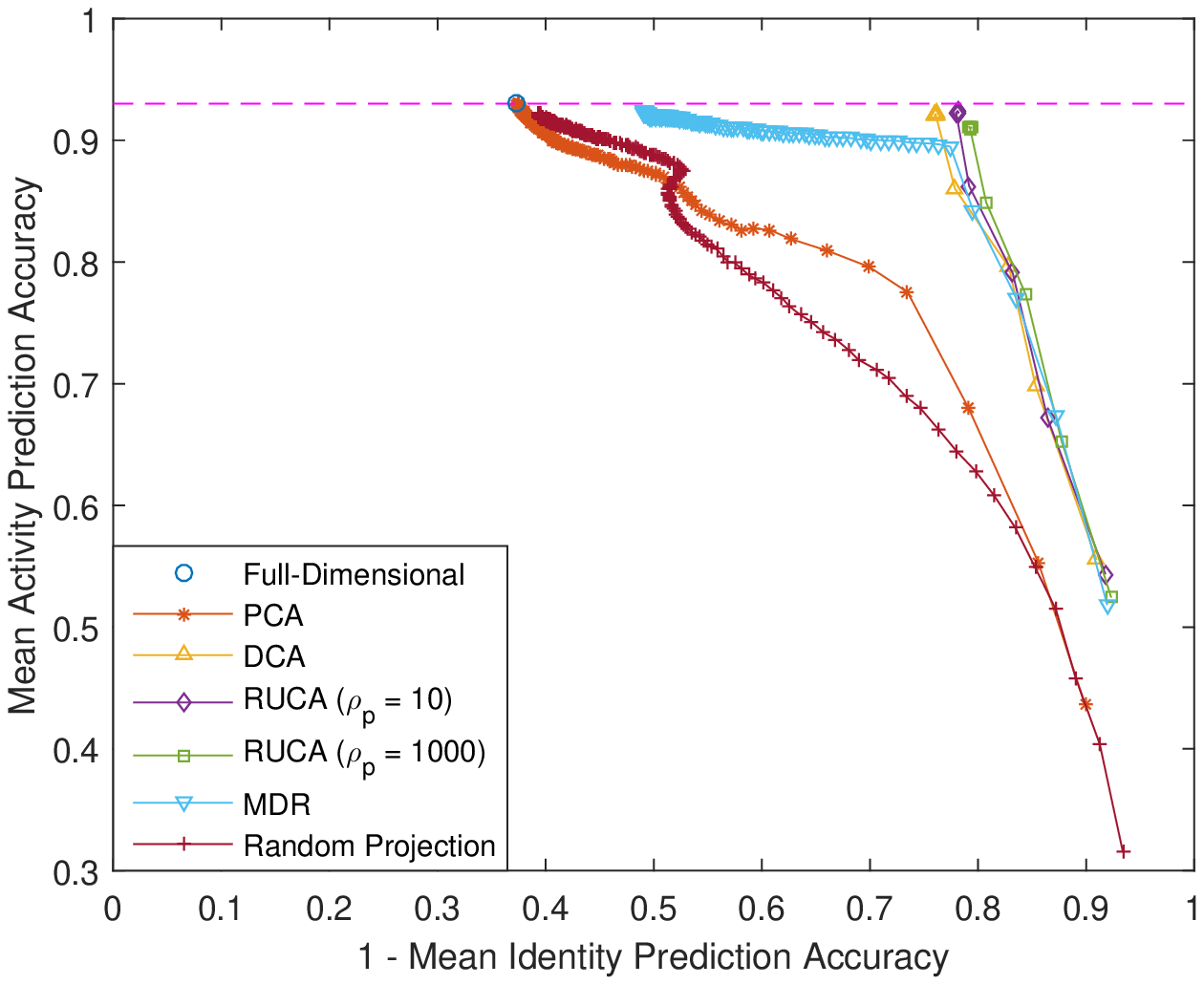}}
  \centerline{(b) HAR (Activity)}\medskip
\end{minipage}
\hfill
\begin{minipage}[b]{0.33\linewidth}
  \centering
  \centerline{\includegraphics[width=6cm]{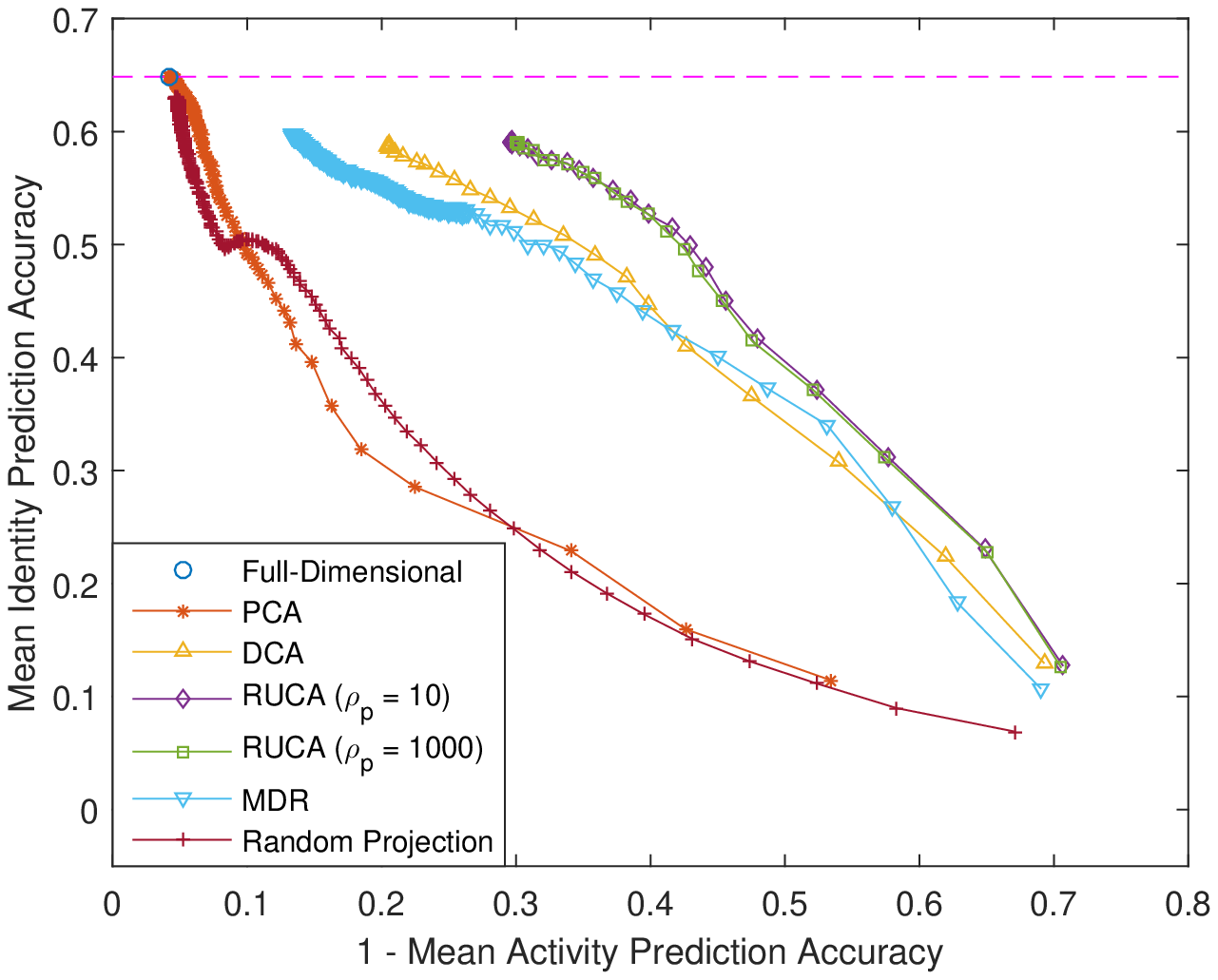}}
  \centerline{(c) HAR (Identity)}\medskip
\end{minipage}
\caption{Utility-Privacy trade-off curves for the datasets showing Mean Utility Prediction Accuracy vs. (1 {-} Mean Privacy Prediction Accuracy). Curves were obtained by increasing the number of components with each projection method. Dashed lines show the utility classification accuracies obtained with full-dimensionality. RUCA outperforms all other methods (a) when $\beta \geq 0.067$ on the Census data set, (b) for all privacy pricings on the HAR data set with Activity as utility, (c) when $\beta \geq 0.226$ on the HAR data set with Identity as utility.}
\label{fig:ROC}
\end{figure*}

To achieve the task outlined above, we need to select a subspace such that separability between classes based on the utility labels $y_i$ is maximized, while separability between classes based on privacy labels $s_i$ is minimized. For utility driven dimension reduction, given the subspace dimension $K$, DCA \cite{SYKungDCA} involves searching for the projection matrix $\W_{DCA} \in \R^{M \times K}$:
\begin{equation}
\W_{DCA} = \underset{\W:\W^T[\Sbar+\rho \I]\W=\I}{\text{arg\ max}}\tr(\W^T \SBu \W)
\label{eq:WDCA}
\end{equation}
where $\tr(\cdot)$ is the trace operator and $\rho \I$ is a small regularization term added for numerical stability. $\Sbar$ is the center adjusted scatter matrix:
\begin{equation}
\Sbar = \Xbar\Xbar^T=\sum_{i=1}^{N}[\x_i-\bmu][\x_i-\bmu]^T
\end{equation}
where $\bmu$ denotes the mean of the samples $\{\x_i\}_{i = 1}^{N}$. $\Sbar$ is divided into two additive parts:
\begin{equation}
\Sbar = \SBu+\SWu
\label{eq:Sbar}
\end{equation}
where $\SBu$ and $\SWu$ are utility between-class and within-class scatter matrices, respectively. These are defined as
\begin{align}
\SBu &= \sum_{c=1}^L N^u_c[\bmu-\bmu^u_c][\bmu-\bmu^u_c]^T \\
\SWu &= \sum_{c=1}^L \sum_{y_i=c} [\x_i-\bmu^u_c][\x_i-\bmu^u_c]^T
\end{align}
where $\bmu^u_c$ is the mean and  $N^u_c$ is the number of samples in utility class $c$, respectively. Privacy between-class scatter matrix $\SBp$ can be defined similarly:
\begin{equation}
\SBp = \sum_{c=1}^P N^p_c[\bmu-\bmu^p_c][\bmu-\bmu^p_c]^T
\end{equation}
where $\bmu^p_c$ is the mean and  $N^p_c$ is the number of samples in privacy class $c$, respectively. 

Optimal solution to the problem given in Equation \ref{eq:WDCA} remains the same when $\Sbar$ is replaced with $\SWu$ due to the relationship given in Equation \ref{eq:Sbar}. Even though Equation \ref{eq:WDCA} applies more restrictive orthonormality constraints to the columns of the projection matrix $\W$, the subspace spanned by these columns constitutes an optimal solution for \textit{Multiclass Discriminant Analysis} (MDA) criterion \cite{DudaPatternBook} (with an additional regularization term $\rho \I$):
\begin{equation}
MDA=\frac{\det(\W^T \SBu \W)}{\det(\W^T (\SWu+\rho \I) \W)}
\end{equation}
where $\det(\cdot)$ is the determinant operator.

In addition, an optimal solution to both of these problems can be derived from the first $K$ principal generalized eigenvectors of  the matrix pencil $(\SBu,\Sbar+\rho \I)$ \cite{GHGolub}.

\textit{Multiclass Discriminant Ratio} (MDR) is a natural extension to MDA criterion for the case where there are two conflicting goals: To maximize separability for a utility classification problem and to minimize separability for a privacy classification problem \cite{Kostas}. It is defined as:
\begin{equation}
MDR = \frac{\det(\W^T \SBu \W)}{\det(\W^T (\SBp+\rho \I) \W)}
\end{equation}

Analogous to DCA and MDA, an optimal solution to MDR can be derived from the first $K$ principal generalized eigenvectors of the matrix pencil $(\SBu,\SBp+ \rho \I)$. Thus DCA and MDR, barring an orthonormality constraint on the columns of the projection matrix, are very similar and can both be solved via Generalized Eigenvalue Decomposition. 

We shall add additional parameters to DCA to obtain a compromise between DCA and MDR, which we will call \textit{Ratio Utility and Cost Analysis} (RUCA):
\begin{equation}
\W_{RUCA} = \underset{\W:\W^T[\mathbf{S}_{RUCA}+\rho \I]\W=\I}{\text{arg\ max}}\tr(\W^T \SBu \W) \\
\end{equation} 
where $\mathbf{S}_{RUCA}$ is a privacy-regularized scatter matrix:
\begin{equation}
\mathbf{S}_{RUCA}=\Sbar+\rho_{p} \SBp
\end{equation}
where $\rho_p$ is a privacy parameter different from $\rho$. 

Note that when $\rho_p$ is 0, this projection method becomes DCA and when $\rho_p$ is very large, it becomes MDR as the term $\rho_p \SBp$ dominates. By varying $\rho_p$, it is possible to establish a more favorable trade-off between utility and privacy than MDR. Additionally, RUCA can be generalized to multiple privacy classifications by including multiple between-class scatter matrices in the regularization:
\begin{equation}
\mathbf{S}_{RUCA}=\Sbar+\sum_{i}^{} {\rho_{p}}_i {\SBp}_i
\end{equation}

Finally, an optimal solution to RUCA can be derived from the first $K$ principal generalized eigenvectors of the matrix pencil $(\SBu,\mathbf{S}_{RUCA}+\rho \I)$. In other words, columns of the projection matrix $\W$ correspond to $K$ largest eigenvalues $\lambda_i$ satisfying the following relationship:
\begin{equation}
\SBu \w_i = \lambda_i (\mathbf{S}_{RUCA}+\rho \I)\w_i
\label{eq:eigen}
\end{equation}

In all the subspace optimization techniques described above, the left hand side of the characteristic equation remains the same as in Equation \ref{eq:eigen}. Due to the fact that rank of $\SBu$ is at most $L-1$, there are at most $L-1$ non-zero eigenvalues associated with the generalized eigenvalue decompositions. In practice, another small regularization term $\rho' \I$ may be added to $\SBu$ to make it full rank, which will allow users to rank the columns of $\W$ in cases where $K \geq L$. As columns corresponding to eigenvalues ranking $L$ or lower don't normally contribute to our criteria, they are expected to have little contribution to the effectiveness of utility classification.

\section{Experimental Results}
\label{sec:eval}
\subsection{Data Sets}

\begin{table}[t]
\centering
\caption{Mean Accuracy Percentages with $K=1$, Income classification being utility, and Marital Status and Gender classifications being privacy on the Census data set. $\rho_p$ is the privacy parameter for Marital Status classification. RUCA outperforms all other methods when $\beta \geq 0.073$.}
\begin{tabularx}{\linewidth}{|l|Y|Y|Y|Y|} \hline
Projection Method & Income & Marital Status & Gender \\ \hline
Random Projection \cite{KunLiu} & 81.07 & 40.22 & 53.63 \\ 
MDR \cite{Kostas} & 81.93 & 36.64 & 54.40 \\ 
RUCA ($\rho_p=16$) & 82.99 & \textbf{35.87} & 51.21 \\ 
RUCA ($\rho_p=8$) & 83.30 & 37.05 & 51.17 \\ 
RUCA ($\rho_p=4$) & 84.07 & 40.40 & 51.31 \\ 
RUCA ($\rho_p=2$) & 84.97 & 45.80 & 51.61 \\ 
RUCA ($\rho_p=1$) & 85.67 & 50.66 & 52.07 \\ 
DCA \cite{SYKungDCA} & \textbf{86.24} & 58.41 & 52.49 \\ 
PCA & 81.06 &  38.20 & 55.59 \\ 
Full-Dimensional & 86.91 & 81.78 & 75.63 \\ \hline
\end{tabularx}
\label{table:Census}
\end{table}

We have tested our approach with multiple applications on Census (Adult) and Human Activity Recognition (HAR) \cite{HAR} data sets, both of which are available at UCI Machine Learning Repository \cite{UCI}. For the Census data set we used Income as the utility classification where we try to classify an individual as with high- or low-income, parallel with the original purpose of the data set. Privacy classifications were chosen as Marital Status and Gender, both of which were given as categorical features in the original data. We grouped `Married-civ-spouse', `Married-spouse-absent' and `Married-AF-spouse' into a single category called `Married'. `Divorced', `Separated' and `Widowed' were grouped into a single category called `Used to be Married'. We left the `Never Married' category as is. 

We first removed the samples with missing features in the data set and randomly sampled the rest of the training and testing sets (separately) in order to create two sets in which all privacy classes have equal number of samples, i.e. numbers of males and females were equal in our training and testing sets, and so were the number of samples categorized as 'Married', 'Never Married' and 'Used to be Married'. All categorical features were turned to numerical ones via binary encoding, as we determined it to yield higher classification accuracies than one-hot encoding with this data. After these operations we had 10086 samples remaining in the training set and 4962 samples remaining in the testing set with 29 features.

In HAR data set, we had Activity and Identity as labels available to us, either of which can be utility or privacy based on the application. Therefore we tested for both cases. Activity had 6 types of labels: `Walking', `Walking Upstairs', `Walking Downstairs', `Sitting', `Standing' and `Laying'. Identity, on the other hand, had 21 types of labels based on the individuals who contributed to the data. 

Training and testing sets of the HAR data set consist of samples contributed by two disjoint sets of users. Therefore we extracted testing sets for Identity classification by randomly picking samples from the original training set. When Activity classification was chosen as utility, we tested Activity classification accuracy on the original testing set and Identity classification accuracy on the extracted testing set. The numbers of training, privacy testing and utility testing samples were 4011, 1890 and 2947, respectively, with 561 features. 

When Identity classification was chosen as utility, we tested both Identity and Activity classification accuracies on the same testing set, which was extracted from the original training set.  The numbers of training and testing samples were 4026 and 1890 respectively with 561 features. As with the Census data, we kept the number of samples in all privacy classes equal in all sets.

\subsection{Results}

\begin{table}[t]
\centering
\caption{Mean Accuracy Percentages with $K=5$, Activity classification being utility, and Identity classification being privacy on the HAR data set. $\rho_p$ is the privacy parameter for Identity classification. RUCA outperforms all other methods when $\beta \leq 5.621$.}
\begin{tabularx}{\linewidth}{|l|Y|Y|Y|Y|} \hline
Projection Method & Activity & Identity \\ \hline
Random Projection \cite{KunLiu} & 56.14 & \textbf{14.73} \\ 
MDR \cite{Kostas} & 89.68 & 22.53 \\ 
RUCA ($\rho_p=1000$) & 90.99 & 20.93 \\ 
RUCA ($\rho_p=100$) & 91.34 & 21.20 \\ 
RUCA ($\rho_p=10$) & 92.24 & 21.98 \\ 
RUCA ($\rho_p=1$) & \textbf{92.37} & 22.98 \\ 
DCA \cite{SYKungDCA} & 92.07 & 23.95 \\ 
PCA & 79.61 & 30.20 \\ 
Full-Dimensional & 93.01 & 62.64 \\ \hline
\end{tabularx}
\label{table:HARa}
\end{table}

All our experiments were performed using RBF SVM on the original and projected data. Training and testing sets were separated as described in the last section before the experiments commenced. With the Census data set we performed 50 iterations at which we randomly picked 10\% of the training samples. At each iteration and with each projection method, a 5-fold cross-validation grid search was performed to find the best parameters for training utility and privacy classifiers. 

With the HAR data set we performed 50 iterations at which we randomly picked 25\% of the training samples. Once again, optimal parameters for SVM-RBF were determined via 5-fold cross-validation at each iteration. PCA and Random Projection were also included in our experiments for comprehensiveness. 

In order to compare RUCA's performance with other projection methods, we adopt a simple performance criterion:
\begin{equation}
Performance = Acc_{U}+\beta(1-Acc_{P})
\end{equation}
where $Acc_U$ and $Acc_P$ denote the utility and privacy classification accuracies, respectively, and $\beta$ denotes the \textit{Privacy Pricing}. Higher $\beta$ indicates that higher emphasis is placed on privacy, while $\beta=0$ indicates that all the emphasis is placed on utility.

Figure \ref{fig:ROC} displays the utility-privacy trade-off curves obtained by progressively adding more components with each projection method. We stopped adding components as they started contributing predominantly to privacy classification. To obtain the results provided in Tables \ref{table:Census}, \ref{table:HARa} and \ref{table:HARb}, we picked $K=1$, $K=5$ and $K=20$, respectively, because we had $L=2$, $L=6$ and $L=21$ for Income, Activity and Identity classification problems, respectively.

The curves in Figure \ref{fig:ROC}(a) demonstrate a trade-off between utility and privacy as the privacy parameter $\rho_p$ is increased. Even RUCA with a low privacy parameter achieves higher privacy levels than possible with DCA. RUCA with $\rho_p=1$ outperforms PCA and DCA when $\beta \geq 0.067$, whereas RUCA with $\rho_p=4$ outperforms MDR and all remaining methods for all privacy pricings. Based on the trade-off curves in Figures \ref{fig:ROC}(b) and \ref{fig:ROC}(c), RUCA outperforms both DCA and MDR on HAR data for all privacy pricings. Furthermore, RUCA outperforms all other methods in (b) for all privacy pricings and all other methods in (c) when $\rho_p \geq 0.226$. PCA and Random Projection, on the other hand, are seen to under-perform in all plots when the privacy pricing is high. 

By comparing the curves in (b) and (c), it becomes apparent that Identity classification when Activity is private is much harder than Activity classification when Identity is private on HAR data. Steepness of the drops in (b) suggests that more utility performance can be obtained by sacrificing relatively little privacy, which is not the case in (c).

Results with $K=1$ for the Census data set are given in Table \ref{table:Census}. Clearly, DCA alone reduces gender classification accuracy close to random guessing (50\%) by sacrificing less than 1\% (absolute) utility classification accuracy. Accordingly for this application, a nonzero privacy parameter $\rho_{p}$ was only applied to the between-class scatter matrix of Marital Status classification and privacy parameter was kept at 0 for Gender classification. The table demonstrates a clear utility-privacy trade-off as $\rho_p$ is increased, similar to Figure \ref{fig:ROC}(a). RUCA outperforms DCA when $\beta \geq 0.073$ and all other methods for all privacy pricings. Results indicate that a small privacy parameter $\rho_p$ provides significantly better privacy while sacrificing little utility classification performance, whereas with a large $\rho_p$ it is possible to get better utility classification performance for the same privacy classification performance as other methods.

Tables \ref{table:HARa} and \ref{table:HARb} show similar results for HAR data set when Activity classification and Identity classification are chosen as utility, respectively. Utility performance doesn't immediately drop, though privacy classification accuracies decrease as $\rho_p$ is increased. Here RUCA outperforms all other methods for all privacy pricings, except for Random Projection as seen in Table \ref{table:HARa}. Although Random Projection provides better privacy for HAR data set when Activity classification is chosen as utility, it only outperforms RUCA when $\beta \geq 5.621$, i.e. when much higher emphasis is placed on privacy.

\begin{table}[t]
\centering
\caption{Mean Accuracy Percentages with $K=20$, Identity classification being utility, and Activity classification being privacy on the HAR data set. $\rho_p$ is the privacy parameter for Activity classification. RUCA outperforms all other methods for all privacy pricings.}
\begin{tabularx}{\linewidth}{|l|Y|Y|Y|Y|} \hline
Projection Method & Identity & Activity \\ \hline
Random Projection \cite{KunLiu} & 38.47 & 81.72 \\ 
MDR \cite{Kostas} & 52.57 & 73.46 \\ 
RUCA ($\rho_p=1000$) & 59.03 & \textbf{69.81} \\ 
RUCA ($\rho_p=100$) & 59.05 & 69.84 \\ 
RUCA ($\rho_p=10$) & \textbf{59.06} & 70.21 \\ 
RUCA ($\rho_p=1$) & 58.91 & 74.70 \\ 
DCA \cite{SYKungDCA} & 58.52 & 79.41 \\ 
PCA & 50.07 & 90.37 \\ 
Full-Dimensional & 64.85 & 95.77 \\ \hline
\end{tabularx}
\label{table:HARb}
\end{table}

\section{Conclusion}
\label{sec:conc}

We have presented a novel subspace projection method that allows data offered by users in a collaborative learning environment to be used for the intended purpose, with minimal loss of private information. We formulated a new criterion called Ratio Utility and Cost Analysis, which combines utility driven DCA  with privacy emphasized MDR. Our method allows users to define multiple undesirable classifications on their data and achieve better utility for a given level of privacy. Using publicly available Census (Adult) and Human Activity Recognition data sets, we have demonstrated that our approach can provide better classification performance for the intended task for an equally low privacy classification performance when compared with state-of-the-art methods. Future work will include the extension of RUCA to privacy preserving non-linear projections, as well as an optimization method for the privacy parameters.

%

{\bibliographystyle{IEEEbib}
\bibliography{refs}}

\nocite{*}

\end{document}